\documentclass[twoside]{article}

\usepackage[accepted]{aistats2022}

\usepackage{amsmath, amsfonts, bm}
\usepackage{graphicx}
\usepackage{url}
\usepackage{hyperref}

\usepackage[round]{natbib}

\begin{document}

%

%

\twocolumn[

\aistatstitle{Increasing the accuracy and resolution of precipitation forecasts using deep generative models}

\aistatsauthor{Ilan Price \And Stephan Rasp }

\aistatsaddress{ University of Oxford \\ The Alan Turing Institute \And ClimateAI} ]

\begin{abstract}
  Accurately forecasting extreme rainfall is notoriously difficult, but is also ever more crucial for society as climate change increases the frequency of such extremes. Global numerical weather prediction models often fail to capture extremes, and are produced at too low a resolution to be actionable, while regional, high-resolution models are hugely expensive both in computation and labour. In this paper we explore the use of deep generative models to simultaneously correct and downscale (super-resolve) global ensemble forecasts over the Continental US. Specifically, using fine-grained radar observations as our ground truth, we train a conditional Generative Adversarial Network---coined CorrectorGAN---via a custom training procedure and augmented loss function, to produce ensembles of high-resolution, bias-corrected forecasts based on coarse, global precipitation forecasts in addition to other relevant meteorological fields. Our model outperforms an interpolation baseline, as well as super-resolution-only and CNN-based univariate methods, and approaches the performance of an operational regional high-resolution model across an array of established probabilistic metrics. Crucially, CorrectorGAN, once trained, produces predictions in seconds on a single machine. These results raise exciting questions about the necessity of regional models, and whether data-driven downscaling and correction methods can be transferred to data-poor regions that so far have had no access to high-resolution forecasts.
\end{abstract}


\section{INTRODUCTION}
Heavy rainfall is one of the most impactful weather extremes, causing substantial economic losses and physical harm each year across the globe \citep{field2014climate}. As climate change progresses, projections show that precipitation extremes will become more frequent and intense \citep{massondelmotte2007climate}. Early and accurate warnings for extreme precipitation are crucial for limiting the resulting damages. 

Most heavy precipitation is the result of small-scale ($\approx$1km) air motions, for example in thunderstorms. However, weather forecasts in large parts of the globe are based on global numerical weather models that have grid spacings of 10 km or larger \citep{bauer2015quiet}. Consequently, these models do a poor job of resolving extreme precipitation, leading to washed-out forecasts. To avoid this, individual countries are running regional, high-resolution weather models that are better able to represent the phenomena leading to extreme rainfall \citep{mesinger2001limited}. This regional modeling approach provides more accurate forecasts than its global counterpart, but has several drawbacks. First, regional models are very labor intensive to develop and run. National weather centers typically employ dozens of scientists for this task. Second, they are computationally expensive, taking hours on large super-computing clusters. Third, these models still exhibit errors, and they are not directly informed by observations, such as weather radars \citep{saltikoff2019overview}. Lastly, because of the difficulty of running such a model, only wealthy nations are able to maintain such services, leaving a large part of the globe without high-resolution weather forecasts. This is especially problematic because extreme precipitation strongly affects poorer nations \citep{field2014climate}. 

\begin{figure*}
    \centering
    \includegraphics[width=\textwidth]{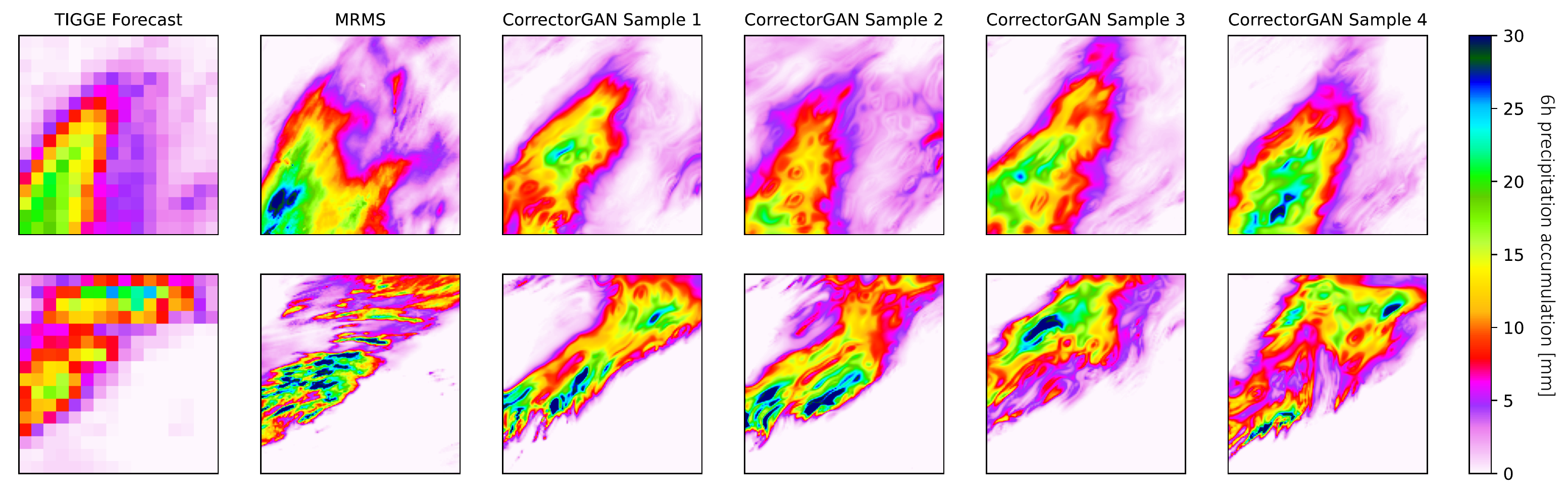}
    \caption{Example forecasts generated by CorrectorGAN, alongside a random ensemble member of the corresponding low-resolution global model forecast (TIGGE) and  high-resolution ground truth (MRMS).}
    \label{fig:figure1}
\end{figure*}

In this paper we explore a data-driven approach to correcting and downscaling (i.e. increasing the resolution of) global model predictions using deep generative modeling. The idea is for the generative model to map directly from the global model's coarse-resolution multi-variable fields to distributions over high-resolution precipitation fields from which the ground truth precipitation observations are a sample. This is a difficult task, as it requires simultaneously 1) correcting biases and errors in low-resolution global model forecasts, 2) super-resolving the global model forecasts into plausible high-resolution fields, and 3) learning the variability in the underlying true distribution so that the generated conditional distributions constitute \textit{reliable} probabilistic forecasts. 

We design a conditional generative adversarial network (cGAN), coined CorrectorGAN, and an associated training regime specifically for this purpose (Section~\ref{sec:Model}), yielding a model which generates an ensemble of plausible predictions. We compare our method to a simple baseline and a state-of-the-art high-resolution regional forecasting system over the Continental United States, using a range of well established metrics (see Section~\ref{sec:Eval}). Our model significantly outperforms the simple baseline and approaches the performance of the high-resolution regional model at a tiny fraction of the cost and effort (Section~\ref{sec:Results} and Figure~\ref{fig:figure1}). 

This is the first study to tackle this problem directly and evaluate against relevant baselines in a setup that closely resembles an operational environment. The results also have several interesting implications. First, they raise questions about the future necessity of expensive regional weather models. If it turns out all such models do is a dynamical downscaling of coarser models, can this be replaced by purely statistical downscaling \citep{palmer2020vision}? Our results represent preliminary evidence to suggest this may be the case.  Second, success of this approach in an area like the United States, which has established radar-observation networks for training and validation, raises a further natural (open) question as to whether such models can be transferred to data-poor regions. If this holds true, it could enable a large jump in forecast quality for some of the regions most affected by heavy precipitation.

\section{RELATED WORK}\label{sec:related}
\noindent \textbf{Post-processing of probabilistic weather forecasts.}
Broadly, methods for probabilistic post-processing of weather forecasts can be divided into those which produce an independent, univariate distribution for each individual pixel (known as univariate methods), and those which include conditional dependencies between pixels (known as multivariate methods).  

When using univariate methods, predictions are made by independently sampling from the distributions of each pixel. Popular such methods include parametric methods, where the parameters of a predetermined distribution are estimated \citep{scheuerer2014probabilistic}, quantile regressions or random forests \citep{taillardat2016calibrated}, or binned/categorical forecasts. \cite{sonderby2020metnet}, for example, parameterized their forecast distribution by binning the possible range of rainfall and estimating per-pixel categorical distributions. A recent deep learning based univariate method developed by \cite{bano2020configuration} uses a CNN-based downscaling architecture to learn a map to a per-pixel Bernoulli-Gamma distribution.

If one is interested in the prediction for a single location, the above-mentioned methods are suitable. However, there are several use cases for which this is not sufficient. One important example is when there are substantial risks associated with large and contiguous regions of rainfall, e.g. a catchment area with the  potential for heavy flooding \citep{ben2016generation}. In such cases, statistical dependencies between forecasts in different pixels are important to model. This motivates multivariate methods.  

One method to convert a univariate into a multivariate forecast is using a global ensemble \textit{post hoc} to sample the post-processed forecast. The Shaake Shuffle \citep{clark2004schaake} and Ensemble Copula Coupling \citep{schefzik2013uncertainty} are examples of this approach. These methods rely on the spatial information of the coarse input being reliable, however---something that cannot always be assumed, especially for precipitation. 

Other studies use a generative modeling approach. Gaussian Random fields have been proposed as a method to generate samples from a spatially coherent distribution \citep{prudden2021stochastic}. \cite{groenke2020climalign} proposes downscaling of climate data using normalizing flows. Similarly, latent neural processes have been used to learn a spatial downscaling of temperature and precipitation \citep{vaughan2021multivariate}. 

\noindent \textbf{GANs in precipitation forecasting}. \cite{leinonen2020stochastic} train a GAN to produce stochastic super-resolved precipitation fields. Their model takes as input a low-resolution time-series sequence of the precipitation in a given area, and outputs time-consistent, high resolution, stochastic samples of the precipitation field over that time. The crucial difference between that work and ours is that their model was trained and evaluated on the downsampled precipitation \textit{observations}. In practice, models do not have access to a low resolution version of the ground truth, and thus the problem of forecast downscaling is not simply one of super-resolution. In this work, the inputs are low resolution \textit{forecasts}, meaning that the model must learn a more general relationship between the inputs and plausible ground truth observations, including both bias correction and resolution refinement.

\cite{ravuri2021skillful} use a GAN for “nowcasting” (very short-term forecasting). Their model generates multiple samples of high-resolution precipitation fields for the upcoming 90 minutes at 5 minute intervals, based on high-resolution radar observations from the the previous 20 minutes. Our work differs in three key ways: 1) the inputs to our model are forecasts, not observations, 2) our outputs are of a finer resolution than the inputs, and 3) ours is not a forecasting model itself, and the forecasts to which it is applied are those with lead times on the order of a few days (at which low resolution forecasts are themselves the most accurate at predicting low-resolution weather situations), rather than a few minutes.

\section{PROBLEM FORMULATION}
Our goal is to approximate the true underlying distribution of precipitation fields at a given time in a given 512 km $\times$ 512 km geographical area (geopatch), using information from the ensemble of low-resolution global model forecasts for that area and its surroundings. Our starting assumption is that this can be modelled as a conditional distribution. To be precise, denote pairs of (low-res forecast, high-res observation) as $(x_i, y_i)$, where $i$ indexes geopatch-time pairs, and $y_i \sim \mathcal{D}_i$, where $\mathcal{D}_i$ is the true distribution over precipitation fields at geopatch-time $i$. We assume that it is possible to model $\mathcal{D}_i$ as the conditional distribution $P(y | x_i)$. Our approach will be to train a conditional GAN in which the generator learns to approximate this conditional distribution, enabling us to sample any number $k$ of high resolution forecasts $\{ \hat{y}_i^{(j)} \}_{j=1}^k$.

\subsection{Data}
We use weather-radar estimates of precipitation obtained by the Multi-Radar/Multi-Sensor (MRMS) system at 4km resolution as the ground truth precipitation values \citep{zhang2016multi} (see supplemental material (SM) for details). We aim to generate samples from the ground truth distribution based on global ensemble forecasts. Here we use the best global ensemble forecast system run by the European Center for Medium-Range Weather Forecasting (ECMWF), available through the open-source THORPEX Interactive Grand Global Ensemble archive \citep{bougeault2010thorpex}, henceforth called TIGGE. As input for our model we use precipitation, 2m temperature, convective available potential energy, and convective inhibition TIGGE fields, at 32km resolution. Precipitation is accumulated over the 6--12h window. We chose this period because it is when TIGGE is most accurate (the first few forecast hours being contaminated by spin-up effects).

We train and validate our model on $16 \times 16$ patches of the TIGGE forecast ensemble, and their corresponding patches of MRMS data ($128 \times 128$), for the whole of 2018 and 2019, and reserve the whole of 2020 for evaluation. To ensure informative evaluation scores, we restrict the evaluation to patches in which at least 90\% of pixels have a radar data quality of $>0.5$ (see SM for details).

Each input patch has 24 channels: the first 10 are TIGGE precipitation forecast ensemble members; the next 10 are TIGGE total column water forecast ensemble members; the next 3 are TIGGE 2m temperature, convective available potential energy, and convective inhibition forecasts taken from the deterministic forecast. These additional variables and their ensembles are included to provide additional information on the basis of which the GAN can correct model errors. The final channel is a downsampled version of a larger patch of size $46 \times 46$ of a TIGGE precipitation forecast ensemble member---that is, the central $16 \times 16$ patch, extended on each side by $15$ pixels. This last channel is included to provide the model with some context on the wider precipitation field outside of the prediction target area. 

For training purposes, we pre-process all $(x_i, y_i)$ by first taking a zero-preserving $\log$ transform, and then shifting and scaling the data to lie in [0,1]. However, all evaluation is performed on the raw data absent normalisation and transformation.

\section{MODEL}\label{sec:Model}
We train a GAN with generator and discriminator architectures described below. For diagrams summarizing these architectures, see the SM.
\subsection{Network Architecture}
\noindent \textbf{Generator.} The task of the generator ($G$) can be thought of two distinct sub-tasks: 1) to correct errors in the low-resolution precipitation forecasts, generating a more accurate distribution over low-resolution representations of the precipitation field, and 2) to refine the resolution of those corrected low-resolution forecasts. We thus design our generator with these tasks in mind. Broadly speaking, the goal of the early stages of the network is to produce corrected, stochastic, low-resolution representations of the precipitation field, given an ensemble of forecasts of precipitation and other weather variables, and some surrounding spatial context. The goal of the second stage of the network, is to refine the resolution of these corrected representations into physically and visually plausible high-resolution forecasts.

\begin{enumerate}
    \item The \textit{Corrector} consists of  convolutional layer followed by two residual blocks, with output channels numbering 64, 128, 255, respectively, and with ReLU activations. A noise sample $z~\sim~\mathcal{N}(0,I_{16\times16})$ for each input is concatenated to the output of the second residual block as the $256^{th}$ channel. The forward pass continues with three more residual blocks, all with 256 output channels and ReLU activations. The output of the final of these residual blocks is then fed into the `super-resolver' (see below), but is also fed into a convolutional layer with a single output channel, whose output we denote as $g(x,z)$, which will be used as a proxy corrected low-resolution forecast, the error of which will be jointly minimised during optimisation along with the GAN loss and other regularisers.

    \item The \textit{Super-resolver} consists of 4 residual blocks with output channels numbering 256, 128, 64, and 32 respectively and leaky-ReLU activations, interspersed by 3 bi-linear upsampling blocks, increasing the resolution from $16 \times 16$ to $128 \times 128$. The output is then passed through a convolutional layer with 1 output channel, and finally through a sigmoid activation, producing the high-res prediction. 
    
\end{enumerate}

\noindent \textbf{Discriminator.} Our discriminator is modelled after that used by \cite{leinonen2020stochastic}. The general idea is to evaluate both whether a given high-resolution precipitation field appears plausible in its own right, and, additionally, whether it represents the ground truth, given the supplied low-resolution forecasts and context.

The high resolution observation  $y$ (or generated prediction $G(x,z)$), and model-forecasts $x$, are first processed independently, each through a convolutional layer followed by three residual blocks, with 32, 64, 128, and 256 output channels respectively, producing intermediate representations $h_1$ and $h_2$. A stride of 2 is used in residual blocks processing the high-resolution input, so that $h_1, h_2 \in \mathbb{R}^{256 \times 16\times 16}$. Next, $h_1$ and $h_2$ are concatenated along the channel dimension, and processed by a further residual block with 256 output channels, the output of which undergoes average pooling, resulting in a vector $\hat{h}_2 \in \mathbb{R}^{256}$. $h_1$ is also processed independently by another residual block with 256 output channels, and average-pooled, producing $\hat{h}_1 \in \mathbb{R}^{256}$. $\hat{h}_2$ and $\hat{h}_1$ are then concatenated, and passed through a linear layer of width 256, a leaky-ReLU activation, and a final linear layer with a scalar output.

\subsection{Loss functions and training}
We design a 3-stage training procedure for the generator informed by the the dual objectives of stochastic forecast correction and super-resolution.

\noindent \textbf{Stage 1 (Low-res correction):} We begin by training the \textit{Corrector} block of $G$ to improve the accuracy and skill of low-resolution forecasts. Specifically, we train with noise $z=0$ to minimise the following loss function,
\begin{align}
    L_{\text{Stage1}} = \| (g(x,0) - y_{\text{coarse}}) \odot (y_{\text{coarse}}+1) \|_1 \\ 
    - \gamma_0 \hat{FSS}(g(x,0), y_{\text{coarse}})
\end{align}
where $y_{\text{coarse}}$ is a downsampled version of $y$ to $16 \times 16$ resolution, and $\hat{FSS}$ is an approximation of the Fractional Skill Score, a well-established metric for precipitation forecasts, where the binary grid is replaced by a continuous, sigmoid approximation \citep{ebert2021cira} (see SM for details). In the first term we use a weighting scheme to focus more on areas with higher precipitation, a technique used by \cite{ravuri2021skillful}, and we use a threshold value of $0.5$ (pre-processed data lies in [0,1]) in the $\hat{FSS}$ term, and set $\gamma_0= 0.1$. Minimising this loss trains the \textit{Corrector} to correct for errors in spatial distribution based on the patterns identified across its multiple ensemble forecast inputs together with the additional contextual variables.

\noindent \textbf{Stage 2 (High-res pre-training):} Next, we continue training the generator, adding an $L1$ loss on the high-resolution output of $G$, as pre-training for the super-resolver block. This is again done with noise $z=0$, and we maintain the low-resolution $L1$ error term, but drop the $\hat{FSS}$ term, resulting in 
\begin{align}
    L_{\text{Stage2}} = \| (g(x,0) - y_{\text{coarse}}) \odot (y_{\text{coarse}}+1) \|_1  \\
    + \| (G(x,0) - y) \odot (y+1) \|_1
\end{align}

\noindent \textbf{Stage 3 (GAN training):}
Finally, we train the full GAN, to solve
\begin{align}
    \min_{\theta_D} \mathbb{E}_{x,y,z} [L_D(x,y,z,\theta_D)], \\
    \min_{\theta_G} \mathbb{E}_{x,z} [L_G(x,z,\theta_G)],
\end{align}
where $L_D$ and $L_G$ are the loss functions defined for the discriminator and generator respectively, and $\theta_{(\cdot)}$ are the respective network parameters. We use the standard Wasserstein loss with gradient penalty \citep{gulrajani2017improved} for $D$,
\begin{align*}
    L_D & = D(x, G(x,z)) - D(x,y) + \lambda (\| \nabla_{\tilde{y}}D(x, \tilde{y})\|_2 - 1)^2,
\end{align*}
where
\begin{align*}
    \tilde{y} & = \epsilon y + (1-\epsilon)G(x,z), \quad \epsilon \sim U(0,1),
\end{align*}

For our generator, we use a Wasserstein loss, with two additional loss terms, 
\begin{align}
    L_G = \mathbb{E}_{x,y} [ \mathbb{E}_z [-D(x, G(x,z))] + \gamma_1 L_{LR} + \gamma_2 L_{HR}].
\end{align}
The additional loss terms are defined as follows,
\begin{align}   
    L_{LR} & = \| (\mathbb{E}_z [g(x,z)] - y_{\text{coarse}}) \odot (y_{\text{coarse}}+1) \|_1, \\
    L_{HR} & = \| (\mathbb{E}_z [G(x,z)] - y) \odot (y+1) \|_1.
\end{align}
These additional loss terms are similar to those used in the pre-training stage, and similarly encourage spatial overlap and intensity similarity of the corrected proxy low-resolution prediction and the high-resolution output with their corresponding ground truths. However, since we aim to model a \textit{distribution} of possible precipitation fields, we follow \cite{ravuri2021skillful} and compare the mean of an ensemble of generated predictions with the ground truth, rather than looking at the error of each ensemble member. 

For full training details and hyperparameters, see SM.

\section{EVALUATION}\label{sec:Eval}
\subsection{Metrics}\label{sec:metrics}
Evaluating probabilistic precipitation forecasts is non-trivial. This is primarily for two reasons: first, forecast distributions tend to be highly non-Gaussian and intermittent, and second, different end users might be interested in only a certain aspect of a forecast, e.g. only extreme precipitation. As a result, no single metric can capture the full spectrum of skill. Here we choose a set of metrics based on the proposition that a good probabilistic forecast should be reliable and sharp \citep{gneiting2007probabilistic}. Reliability is a key property requiring the forecast distribution to be, in a statistical average, a true representation of the actual forecast uncertainty. For example, considering all cases in which rain was forecast with 30\% probability, rain should have actually occurred in 30\% of these cases (see Reliability Diagram below). 

However, reliability alone is not sufficient for a useful forecast. For example, simply predicting the climatological average
is a very reliable but not very useful forecast. A useful forecast also has to be sharp, that is, its distribution should be as narrow as possible while still being reliable. We test jointly for reliability and sharpness using the Brier Score and the Continuous Ranked Probability Score (CRPS). 

\begin{table*}[!htbp]
\centering
\caption{Probabilistic metrics described in Section \ref{sec:metrics} }
\vspace{2mm}
\begin{tabular}{lccccc}
Model              & CRPS  & \multicolumn{4}{c}{Brier Score} \\ \hline
                  &       & 1mm     & 5mm    & 10mm  & 30mm  \\ \cline{3-6} 
CorrectorGAN       & 0.574 & 0.06    & 0.034  & 0.02 & 0.0024  \\
HREF               & 0.562 & 0.059   & 0.032  & 0.019   & 0.0026 \\
TIGGE Interp. & 0.605 & 0.064   & 0.035  & 0.021  & 0.0025 \\
Pure-SR GAN         & 0.61 & 0.063   & 0.036  & 0.021 & 0.0024 \\
BG-CNN              & 0.62 & 0.06   & 0.035  & 0.021  & 0.0032 \\
\hline

\end{tabular}
\label{tab:metrics}
\end{table*}

\noindent \textbf{Reliability Diagram:} Reliability diagrams \citep{wilks2011statistical} plot the conditional distribution of the observations, given the forecast probability, against the forecast probability of binarized precipitation events. Specifically, we choose a precipitation threshold, in our case 1 mm (``light rain"),  5 mm (``moderate rain"), or 10 mm (``heavy rain"), and binarize the forecasts. We estimate the forecast probability from the 10 ensemble members.
Next, we bin the range [0,1] into bins of width 0.2, and check: out of all instances where the event probability fell in a given bin, in what proportion of those instances did the event actually occur? A perfectly reliable forecast lies on the $y=x$ line.

\noindent \textbf{Brier Score:} The Brier Score \citep{wilks2011statistical} also works on thresholded forecast fields. It is defined as 
\begin{align*}
    \text{BS} = \frac{1}{n} \sum^n_{k=1} (y_k - o_k)^2
\end{align*}
where $y_k$ is the event probability, $o_k$ is the corresponding observation, either 0 or 1, and $n$ is the number of samples, in our case all pixels for all forecast times in the test set. Lower scores are better. 

\noindent \textbf{Continuous Ranked Probability Score (CRPS):} The CRPS \citep{wilks2011statistical} is equivalent to an integral of the Brier Score over all thresholds and is defined, for a single point, as 
\begin{align*}
    \text{CRPS} = \int_{-\infty}^\infty [F(y) - F_o(y)]^2 dy,
\end{align*}
where $F(y)$ is the prediction/forecast CDF of the predictand $y$, and,
\begin{align*}
    F_o(y) = \begin{cases}
      0 & y<\text{observed value}\\
      1 & y\geq \text{observed value}.
    \end{cases} 
\end{align*}
$F(y)$ is approximated using the ensemble of forecasts generated by a given model.

\subsection{Baselines}
As important as choosing good metrics is choosing meaningful baselines. Here we choose four baselines in particular. First, as a `lower bound' we bi-linearly interpolate 10 ensemble members of the coarse-resolution global forecast TIGGE to 4 km. Second, we compare against the top performing CNN-based downscaling architecture (denoted BG-CNN) from \citep{bano2020configuration}. Though this is a univariate method, and thus suffers the drawbacks thereof discussed in Section \ref{sec:related}, in particular the lack of spatial covariances, it was nonetheless shown by \cite{groenke2020climalign} to be superior to their proposed normalising flows method on precipitation. Third, to assess the importance of bias-correction, we train a version of our GAN to perform pure super-resolution (trained with 0 noise and low-resolution ground-truth inputs) which we denote Pure-SR GAN. We created an ensemble for evaluation by super-resolving each member of the TIGGE ensemble independently. Finally and most importantly, we evaluate against the High-Resolution Ensemble Forecast (HREF) system \citep{roberts2019href}, an ensemble of storm-resolving regional forecast models run for the Continental US. HREF combines 5 distinct forecast models, each with additional lagged member, i.e. using the forecast initialized 12 hours previously, making it a 10 member ensemble. For further details of the baselines refer to the SM. 

Regional models like HREF are run at high computation and labor costs, only possible for small (wealthy) regions of the world, whereas our forecasts are, once trained, essentially free and can potentially be applied globally. Thus if we can achieve results which are close to HREF (and significantly better than the interpolated TIGGE baseline), this will be a significant step forward.

To ensure fair comparison, we also evaluate CorrectorGAN using an ensemble of only 10 predictions. Note, however, that it is possible to create as large an ensemble as one wants with our model. 

\section{RESULTS}\label{sec:Results}

\begin{figure}[h!]
    \centering
    \includegraphics[width = 0.4\textwidth]{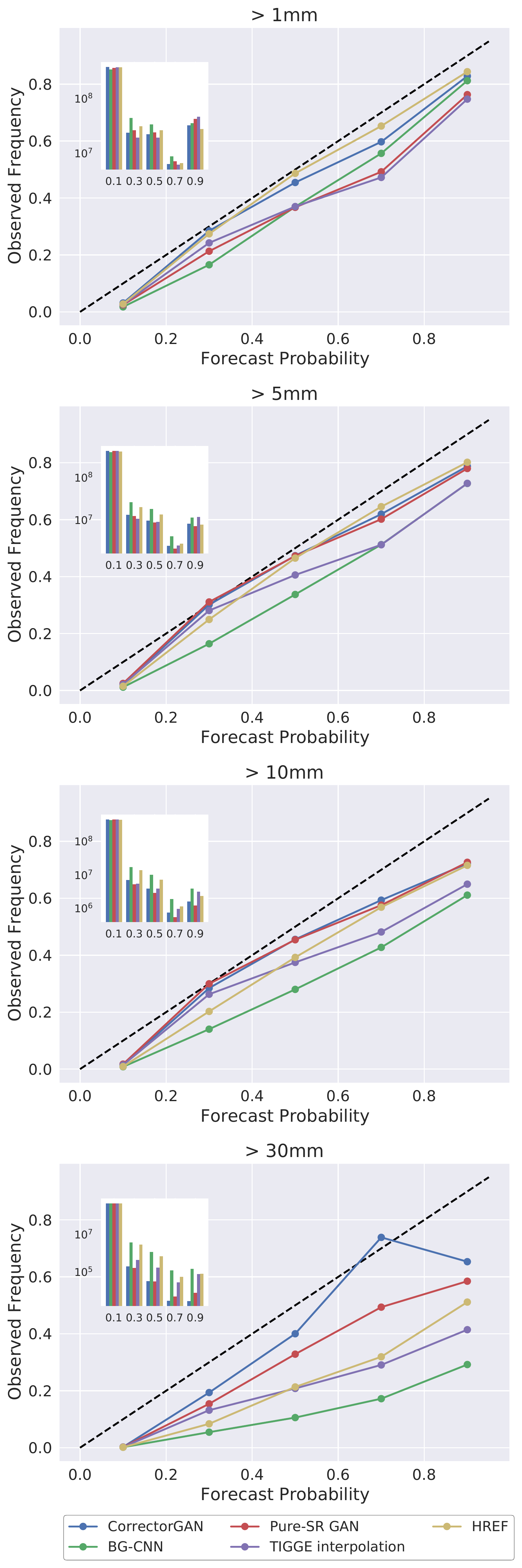}
    \caption{Reliability diagrams comparing the different models, for 1mm, 5mm, 10mm, and 30mm precipitation thresholds. The insets show the number of predictions in each bin for each method (on a log scale).}
    \label{fig:reliability}
\end{figure}

Table~\ref{tab:metrics} shows the key probabilistic metrics described above. CorrectorGAN matches or outperforms the interpolated TIGGE baseline, Pure-SR GAN, and BG-CNN, on both CRPS as well as Brier Scores at all chosen thresholds (1mm, 5mm, 10mm, and 30mm -- the latter constituting approx. the 99.7th percentile).  HREF tends to be slightly better than CorrectorGAN. For CRPS, CorrectorGAN is substantially closer to HREF than any of the other methods, and much closer to HREF than Pure-SR GAN and the TIGGE interpolation on the 1mm Brier Score. For larger thresholds the differences in the Brier Score between the models are quite small. 

\begin{figure*}[!h]
    \centering
    \vspace{0.3cm}
    \includegraphics[width=\textwidth]{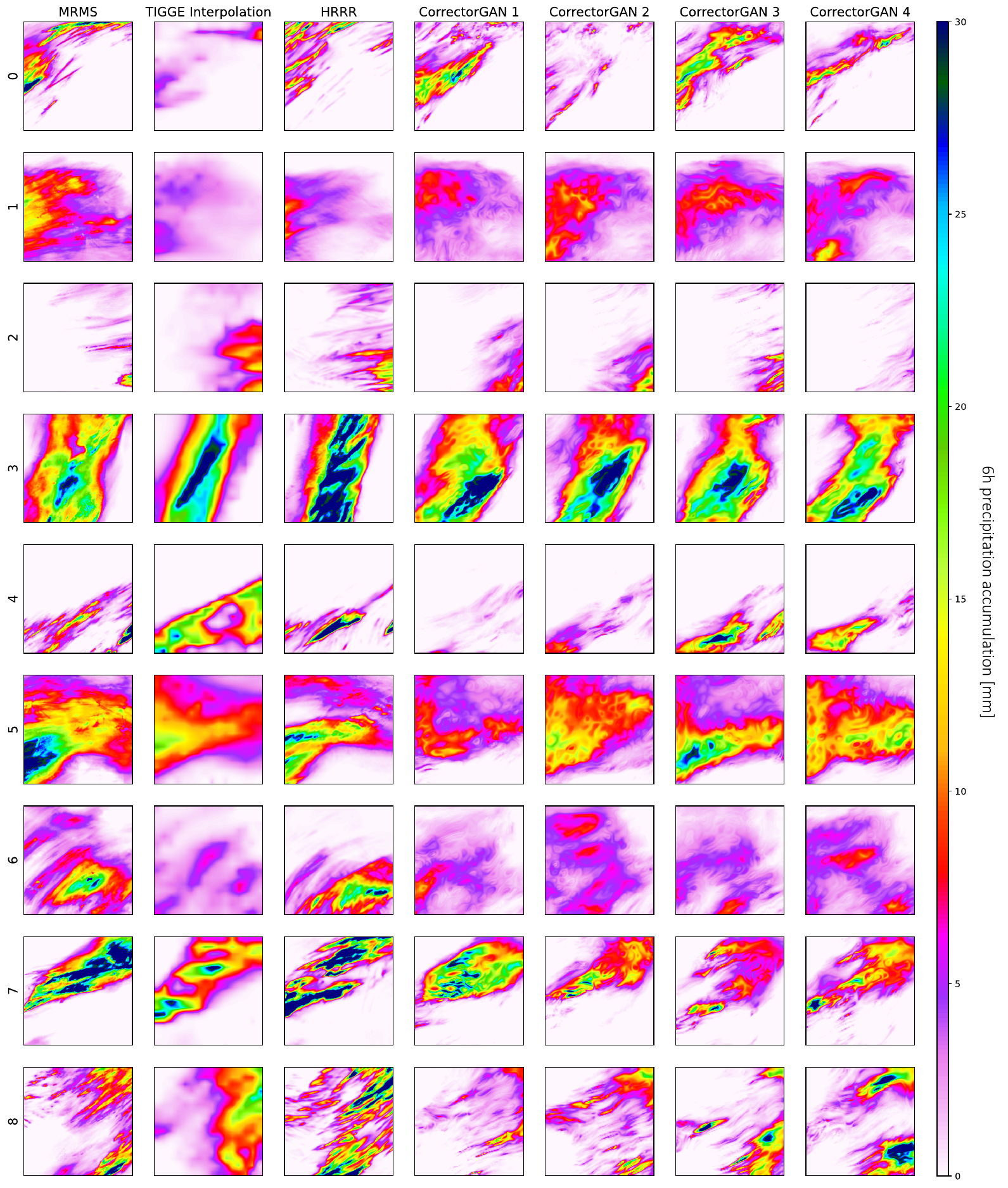}
    \caption{Sample forecasts by different methods alongside the ground truth (MRMS): a randomly selected TIGGE Interpolation ensemble member, one ensemble member of HREF called HRRR, and 4 sample forecasts generated by CorrectorGAN.}
    \label{fig:samples}
\end{figure*}

Figure~\ref{fig:reliability} shows the reliability diagram for all three models, for 1mm, 5mm, 10mm, and 30mm thresholds. For the 1mm and 5mm diagrams, and much of the 10mm diagram, HREF is much closer to the diagonal than TIGGE, indicating a more reliable forecast. TIGGE shows a slightly flatter slope which is a sign of an overconfident forecast \citep{wilks2011statistical}, i.e. a forecast that does not have enough variance given its average forecast error. This is confirmed by other metrics such as the rank histogram shown in the SM. BG-CNN proves in general to be the least reliable, except for the 1mm case for events forecast with high probability, in which case it is more reliable than TIGGE and Pure-SR GAN. CorrectorGAN manages to correct the TIGGE forecasts to be much more reliable, achieving very similar results to HREF, and even providing a generally more reliable forecast given a 10mm or 30mm threshold, i.e. for extreme precipitation. Pure-SR GAN achieves very similar reliability results to CorrectorGAN at 5mm and 10mm thresholds, but substantially worse reliability at 1mm and 30mm thresholds. 

Finally, consider the sample forecasts shown in Figure~\ref{fig:samples}, where the ground truth (MRMS) is shown alongside a randomly selected TIGGE Interpolation ensemble member, one ensemble member of HREF called HRRR, and 4 sample forecasts generated by CorrectorGAN. The examples included in the figure were selected to illustrate strengths and weaknesses of the CorrectorGAN model and its typical behaviour in regions with substantial rainfall. We have not included examples of patches with little to no rainfall observed, since they are the least visually informative. However, the generated forecasts in those instances behave as desired, predicting little-to-no rain. 

It is apparent that CorrectorGAN does two things. First, it is able, in certain instances, to correct for large-scale biases of TIGGE. This is evident, for example, in row 1, where we see that the selected interpolated TIGGE ensemble member (and all other ensemble members too; see the SM) predicts little rain everywhere except along the left most edge. In contrast, all of the CorrectorGAN realizations predict precipitation more closely resembling the MRMS field. Other examples of this bias correction can be seen in rows 0, 2, and 8. This provides some evidence that through the inclusion of total column water, 2m temperature, convective available potential energy and convective inhibition, the GAN is able to extract information about the probability of rainfall beyond just super-resolving the coarse-resolution inputs.

Second, CorrectorGAN adds detail and small-scale extremes. In many cases, the GAN predictions have significantly higher extreme values than the global TIGGE model, corresponding to those present in the MRMS and HREF fields---see for example rows 0, 2, 5, and 8. This, in combination with the improvements in the metrics, confirms that our model achieves its goal of improving extreme precipitation forecasts of global models, approaching the skill of regional high-resolution models but at much reduced cost and effort.

\section{CONCLUSION AND FUTURE WORK}
In this work we trained a conditional GAN --- coined CorrectorGAN --- to correct and downscale the precipitation forecasts of a global numerical weather model. In contrast to a traditional super-resolution task, for the problem at hand the GAN is required to also correct errors in the coarse forecast. To accomplish this we developed a novel two-stage architecture, in which the coarse precipitation forecast is first mapped to a corrected distribution based on information about the weather situation, and this distribution is then mapped to a distribution of high-resolution, plausible predictions.

We compared our method against an interpolation baseline, a super-resolution GAN, a CNN-based downscaling method, and an operational high-resolution regional weather model. CorrectorGAN matches or outperforms the interpolation baseline, super-resolution GAN, and the CNN across all of the probabilistic evaluation metrics. CorrectorGAN's performance is close to that of the high-resolution model, even outperforming it in terms of reliability for extreme rainfall. In contrast to regional weather models, which are expensive to run, CorrectorGAN is fast and cheap. 

There still is room for improvement for generative downscaling and correction methods. While our predictions mostly show realistic small-scale features, some of the forecasts still look ``wavy" --- see e.g. row 5 in Figure~\ref{fig:samples}. We expect that longer training and/or an improved GAN architecture could avoid these artifacts. Furthermore, the GAN is not always able to correct the large scale errors in the TIGGE forecast, resulting in poorer predictions (e.g. row 6, Figure~\ref{fig:samples}). To minimize the amount this happens, one could investigate incorporating further weather variables as inputs, or finding further ways of feeding in information about the local and surrounding weather situation.

In future work, one could include recent radar observations in the input to the GAN, giving the model more temporal high-resolution context. Additionally, we would like to evaluate our model for more forecast times, i.e. beyond 12 hours, to see whether the improvement in skill persists. One could also include a temporal component in the model using recurrent neural network blocks to create temporally coherent realizations.


Finally, one of the most enticing, thought tentative, conclusions from this study is that it seems possible predict extreme precipitation with a similar skill to a regional weather model, using only data (without expensive physics simulations). An important open question is whether it is possible to use such a model trained over data-rich regions like the US and apply it to other parts of the world that do not have access to high-resolution models or observations. To test this, one could verify our model against radar-observations elsewhere, e.g. over Europe, or against station observations in high-risk regions like West Africa.

\bibliographystyle{apalike}
\bibliography{sample}

\begin{thebibliography}{}

\bibitem[Ba{\~n}o-Medina et~al., 2020]{bano2020configuration}
Ba{\~n}o-Medina, J., Manzanas, R., and Guti{\'e}rrez, J.~M. (2020).
\newblock Configuration and intercomparison of deep learning neural models for
  statistical downscaling.
\newblock {\em Geoscientific Model Development}, 13(4):2109--2124.

\bibitem[Bauer et~al., 2015]{bauer2015quiet}
Bauer, P., Thorpe, A., and Brunet, G. (2015).
\newblock The quiet revolution of numerical weather prediction.
\newblock {\em Nature}, 525(7567):47--55.

\bibitem[Ben~Bouall{\`e}gue et~al., 2016]{ben2016generation}
Ben~Bouall{\`e}gue, Z., Heppelmann, T., Theis, S.~E., and Pinson, P. (2016).
\newblock Generation of scenarios from calibrated ensemble forecasts with a
  dual-ensemble copula-coupling approach.
\newblock {\em Monthly Weather Review}, 144(12):4737--4750.

\bibitem[Bougeault et~al., 2010]{bougeault2010thorpex}
Bougeault, P., Toth, Z., Bishop, C., Brown, B., Burridge, D., Chen, D.~H.,
  Ebert, B., Fuentes, M., Hamill, T.~M., Mylne, K., et~al. (2010).
\newblock The thorpex interactive grand global ensemble.
\newblock {\em Bulletin of the American Meteorological Society},
  91(8):1059--1072.

\bibitem[Clark et~al., 2004]{clark2004schaake}
Clark, M., Gangopadhyay, S., Hay, L., Rajagopalan, B., and Wilby, R. (2004).
\newblock The schaake shuffle: A method for reconstructing space--time
  variability in forecasted precipitation and temperature fields.
\newblock {\em Journal of Hydrometeorology}, 5(1):243--262.

\bibitem[Daniel and Wilks, 2006]{daniel2006statistical}
Daniel, S. and Wilks, W. (2006).
\newblock {\em Statistical Methods in the Atmospheric Schiences.}
\newblock Academic Press.

\bibitem[Ebert-Uphoff et~al., 2021]{ebert2021cira}
Ebert-Uphoff, I., Lagerquist, R., Hilburn, K., Lee, Y., Haynes, K., Stock, J.,
  Kumler, C., and Stewart, J.~Q. (2021).
\newblock Cira guide to custom loss functions for neural networks in
  environmental sciences--version 1.
\newblock {\em arXiv preprint arXiv:2106.09757}.

\bibitem[Field and Barros, 2014]{field2014climate}
Field, C.~B. and Barros, V.~R. (2014).
\newblock {\em Climate change 2014--Impacts, adaptation and vulnerability:
  Regional aspects}.
\newblock Cambridge University Press.

\bibitem[Gneiting et~al., 2007]{gneiting2007probabilistic}
Gneiting, T., Balabdaoui, F., and Raftery, A.~E. (2007).
\newblock Probabilistic forecasts, calibration and sharpness.
\newblock {\em Journal of the Royal Statistical Society: Series B (Statistical
  Methodology)}, 69(2):243--268.

\bibitem[Groenke et~al., 2020]{groenke2020climalign}
Groenke, B., Madaus, L., and Monteleoni, C. (2020).
\newblock Climalign: Unsupervised statistical downscaling of climate variables
  via normalizing flows.
\newblock In {\em Proceedings of the 10th International Conference on Climate
  Informatics}, pages 60--66.

\bibitem[Gulrajani et~al., 2017]{gulrajani2017improved}
Gulrajani, I., Ahmed, F., Arjovsky, M., Dumoulin, V., and Courville, A. (2017).
\newblock Improved training of wasserstein gans.
\newblock {\em arXiv preprint arXiv:1704.00028}.

\bibitem[Leinonen et~al., 2020]{leinonen2020stochastic}
Leinonen, J., Nerini, D., and Berne, A. (2020).
\newblock Stochastic super-resolution for downscaling time-evolving atmospheric
  fields with a generative adversarial network.
\newblock {\em IEEE Transactions on Geoscience and Remote Sensing}.

\bibitem[Masson-Delmotte et~al., 2021]{massondelmotte2007climate}
Masson-Delmotte, V. et~al. (2021).
\newblock Climate change 2021: The physical science basis. summary for
  policymakers.
\newblock {\em Sixth Assessment Report of the Intergovernmental Panel on
  Climate Change}.

\bibitem[Mesinger, 2001]{mesinger2001limited}
Mesinger, F. (2001).
\newblock Limited area modeling: Beginnings, state of the art, outlook.
\newblock In {\em 50th Anniversary of Numerical Weather Prediction
  Commemorative Symposium, Book of Lectures}, pages 91--118. Europ. Meteor.
  Soc.

\bibitem[Palmer, 2020]{palmer2020vision}
Palmer, T. (2020).
\newblock A vision for numerical weather prediction in 2030.
\newblock {\em arXiv preprint arXiv:2007.04830}.

\bibitem[Prudden et~al., 2021]{prudden2021stochastic}
Prudden, R., Robinson, N., Challenor, P., and Everson, R. (2021).
\newblock Stochastic downscaling to chaotic weather regimes using spatially
  conditioned gaussian random fields with adaptive covariance.
\newblock {\em arXiv preprint arXiv:2105.08188}.

\bibitem[Ravuri et~al., 2021]{ravuri2021skillful}
Ravuri, S., Lenc, K., Willson, M., Kangin, D., Lam, R., Mirowski, P.,
  Fitzsimons, M., Athanassiadou, M., Kashem, S., Madge, S., et~al. (2021).
\newblock Skillful precipitation nowcasting using deep generative models of
  radar.
\newblock {\em arXiv preprint arXiv:2104.00954}.

\bibitem[Roberts et~al., 2019]{roberts2019href}
Roberts, B., Gallo, B.~T., Jirak, I.~L., and Clark, A.~J. (2019).
\newblock The high resolution ensemble forecast (href) system: Applications and
  performance for forecasting convective storms.
\newblock {\em Earth and Space Science Open Archive}, page~1.

\bibitem[Roberts, 2008]{roberts2008assessing}
Roberts, N. (2008).
\newblock Assessing the spatial and temporal variation in the skill of
  precipitation forecasts from an nwp model.
\newblock {\em Meteorological Applications: A journal of forecasting, practical
  applications, training techniques and modelling}, 15(1):163--169.

\bibitem[Saltikoff et~al., 2019]{saltikoff2019overview}
Saltikoff, E., Friedrich, K., Soderholm, J., Lengfeld, K., Nelson, B., Becker,
  A., Hollmann, R., Urban, B., Heistermann, M., and Tassone, C. (2019).
\newblock An overview of using weather radar for climatological studies:
  Successes, challenges, and potential.
\newblock {\em Bulletin of the American Meteorological Society},
  100(9):1739--1752.

\bibitem[Schefzik et~al., 2013]{schefzik2013uncertainty}
Schefzik, R., Thorarinsdottir, T.~L., and Gneiting, T. (2013).
\newblock Uncertainty quantification in complex simulation models using
  ensemble copula coupling.
\newblock {\em Statistical science}, 28(4):616--640.

\bibitem[Scheuerer, 2014]{scheuerer2014probabilistic}
Scheuerer, M. (2014).
\newblock Probabilistic quantitative precipitation forecasting using ensemble
  model output statistics.
\newblock {\em Quarterly Journal of the Royal Meteorological Society},
  140(680):1086--1096.

\bibitem[S{\o}nderby et~al., 2020]{sonderby2020metnet}
S{\o}nderby, C.~K., Espeholt, L., Heek, J., Dehghani, M., Oliver, A., Salimans,
  T., Agrawal, S., Hickey, J., and Kalchbrenner, N. (2020).
\newblock Metnet: A neural weather model for precipitation forecasting.
\newblock {\em arXiv preprint arXiv:2003.12140}.

\bibitem[Taillardat et~al., 2016]{taillardat2016calibrated}
Taillardat, M., Mestre, O., Zamo, M., and Naveau, P. (2016).
\newblock Calibrated ensemble forecasts using quantile regression forests and
  ensemble model output statistics.
\newblock {\em Monthly Weather Review}, 144(6):2375--2393.

\bibitem[Vaughan et~al., 2021]{vaughan2021multivariate}
Vaughan, A., Lane, N.~D., and Herzog, M. (2021).
\newblock Multivariate climate downscaling with latent neural processes.
\newblock {\em Climate Change AI Workshop}.

\bibitem[Wilks, 2011]{wilks2011statistical}
Wilks, D.~S. (2011).
\newblock {\em Statistical methods in the atmospheric sciences}, volume 100.
\newblock Academic press.

\bibitem[Zhang et~al., 2016]{zhang2016multi}
Zhang, J., Howard, K., Langston, C., Kaney, B., Qi, Y., Tang, L., Grams, H.,
  Wang, Y., Cocks, S., Martinaitis, S., et~al. (2016).
\newblock Multi-radar multi-sensor (mrms) quantitative precipitation
  estimation: Initial operating capabilities.
\newblock {\em Bulletin of the American Meteorological Society},
  97(4):621--638.

\end{thebibliography}


\clearpage
\appendix

\thispagestyle{empty}

\onecolumn \makesupplementtitle

\section{CODE}
Code can be found at \href{https://github.com/raspstephan/nwp-downscale}{https://github.com/raspstephan/nwp-downscale}.

\section{ARCHITECTURES}
\begin{figure}[h!]
	\centering
	\includegraphics[width=\textwidth]{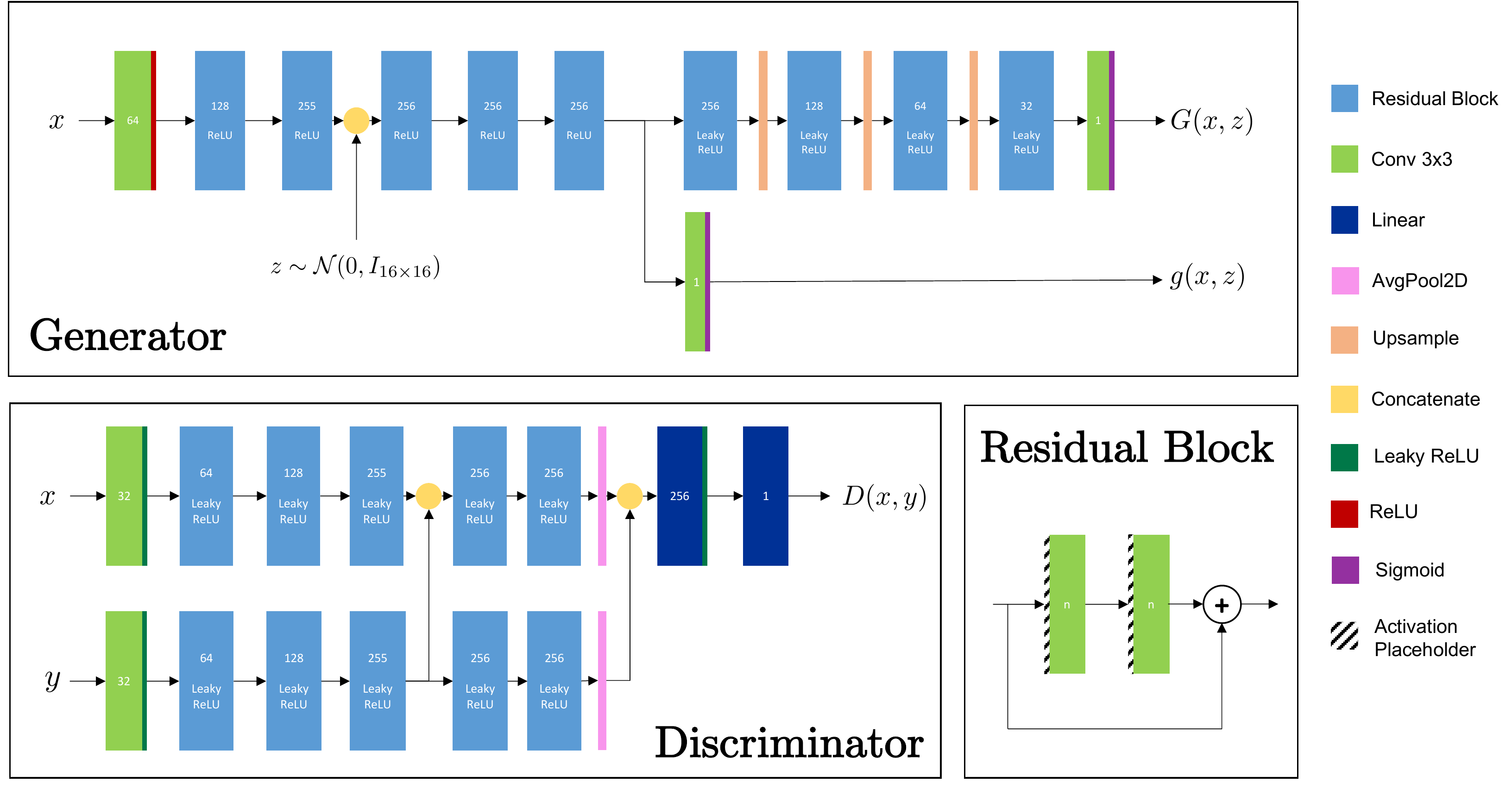}
	\caption{Summary diagrams of the Generator and Discriminator in CorectorGAN. Layers show the number of output channels (or rows in the linear layer case), and their respective internal non-linearities if applicable.}
	\label{fig:arch}
\end{figure}

Figure \ref{fig:arch} shows diagrams illustrating the CorrectorGAN architectures.

\section{TRAINING HYPERPARAMETERS}

\noindent \textbf{Stage 1.} We use Adam with learning rate 5e-05, $\beta_1 = 0$ and $\beta_2=0.9$, and train for 5 epochs, with batch size 128.

\noindent \textbf{Stage 2.} We use Adam with learning rate 5e-05, $\beta_1 = 0$ and $\beta_2=0.9$, and train for 7 epochs, with batch size 128.

\noindent \textbf{Stage 3.} We use Adam with learning rate 5e-05, $\beta_1 = 0$ and $\beta_2=0.9$, for both generator and discriminator, and train for 35 epochs, with batch size 256 and select the final model based on validation CRPS. We set $\gamma_1=20$ and $\gamma_2=20$, $\lambda=10$, and use 6 ensemble members when computing the expectation in the $L_{LR}$ and $L_{HR}$ loss terms.  We train the discriminator for 5 steps for every 1 training step of the generator.

We train on 4 Tesla T4 GPUs using mixed precision.

\section{DATA}

One general note is that we use km throughout the paper, even though all the data is in fact in degrees, since most readers are more familiar with km measurements. For this we use the approximate conversion $0.01^\circ = 1$km.

\subsection{Weighted sampling during training}
Since a large part of the patches have no or very little precipitation, we balance the dataset by preferentially sampling patches with more precipitation. Specifically, we compute, for each patch, the fraction of grid point with precipitation larger than 0.025 mm, denoted by $frac$. To compute the sample weight $w$ we compute: 
\begin{equation}
    w = w_\mathrm{min} + (1 - (frac - 1)^a) * (w_\mathrm{max} - w_\mathrm{min})
\end{equation}
where $w_\mathrm{min} = 0.02$ and $w_\mathrm{max} = 0.4$ are the enforced minimum and maximum weights and the exponent $a = 4$.

\subsection{Input patch}
\begin{figure}[h!]
	\centering
	\includegraphics[width=0.8\textwidth]{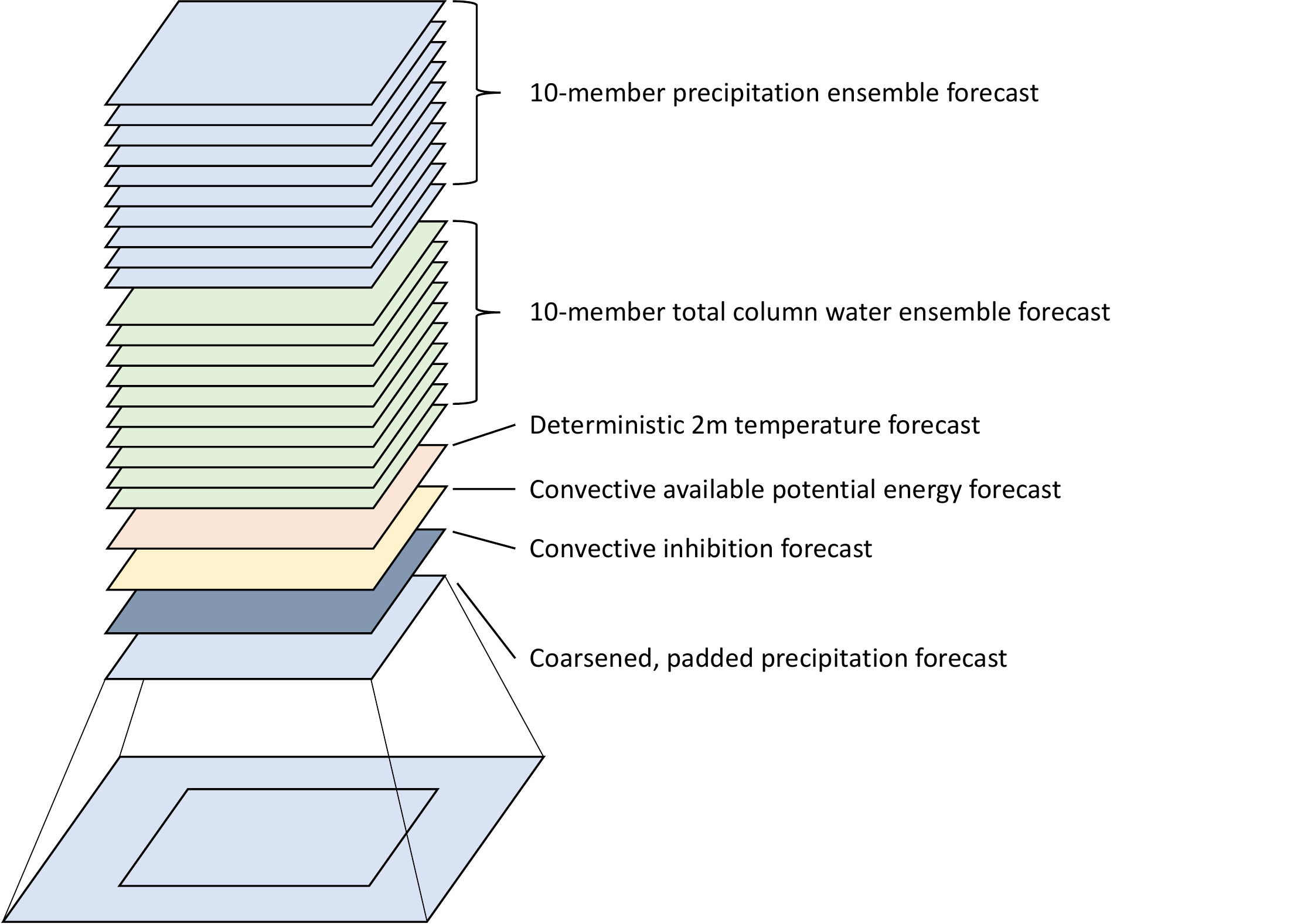}
	\caption{Input patch}
	\label{fig:patch}
\end{figure}

Figure \ref{fig:patch} illustrates the channel breakdown of the inputs supplied to the generator. 

\subsection{TIGGE Ensemble}

TIGGE provides ensemble forecasts from different forecasting centers. We only use the ECMWF Integrated Forecast System operational ensemble and download data from \url{https://apps.ecmwf.int/datasets/data/tigge}. The raw data comes at $\sim 0.13^\circ$ resolution, which we regrid bi-linearly to $0.32^\circ$ resolution. Forecasts are initialized at 00 and 12UTC each day.

\subsection{Multi-Radar/Multi-Sensor (MRMS) Observations}

MRMS data is downloaded from an archive provided by Iowa State University, at \url{https://mtarchive.geol.iastate.edu/}. In particular we download the radar-only 6h quantitative precipitation estimates of MRMS. For details, see \cite{zhang2016multi}. Data originally is provided at $0.01^\circ$ resolution, which we regrid bi-linearly to $0.04^\circ$ resolution.

\begin{figure}[h!]
	\centering
	\includegraphics[width=0.5\textwidth]{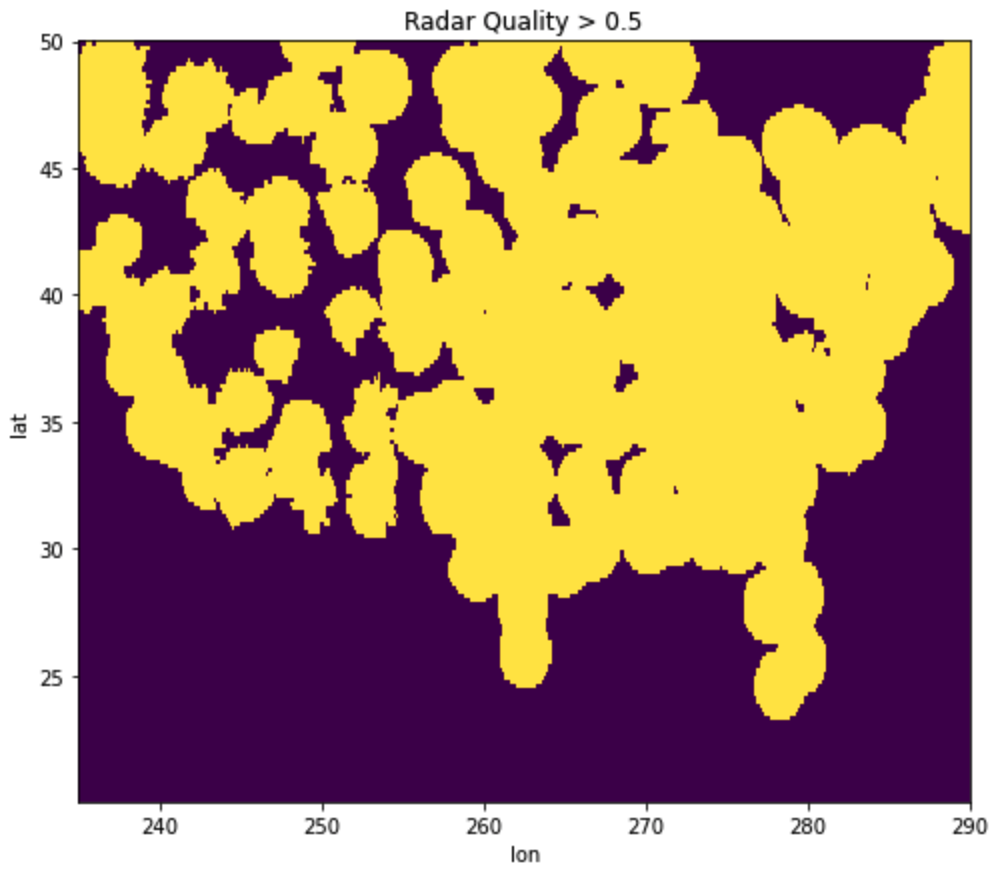}
	\caption{Radar quality mask $> 0.5$}
	\label{fig:rq}
\end{figure}

\noindent \textbf{Radar Quality.} In addition, MRMS data comes with a radar quality index, ranging from 0 to 1. We restrict our evaluation to only those patches with a radar quality of $> 0.5$ in $>90\%$ of their pixels. Figure~\ref{fig:rq} shows the coverage resulting from a $>0.5$ quality mask.

\subsection{HREF High-resolution Ensemble}

HREF data is downloaded from the server of the National Severe Storms Laboratory at \url{https://data.nssl.noaa.gov/thredds/catalog/FRDD/HREF.html}. Here the data is provided for each of the 5 models separately. For each we download the 00 and 12UTC initialization times. We then create a 10 member ensemble by stacking the 5 original models in addition to a lagged forecast (i.e. the forecast initialized 12h earlier) for each model. Note that this is slightly different from the operational HREF version which uses a 6h lag for one of the models, the High-Resolution Rapid Refresh model (HRRR). 

Great care has to be taken in combining the different models because they report precipitation in different formats. Some models report total precipitation accumulation, i.e. the file for 12h forecast lead time contains the precipitation amount from 0 to 12h. To get the 6-12h accumulation, one simply subtracts the 6h values from the 12h values. Other models, however, report only the one hour accumulation in each file. For these models one has to download all files from 7 to 12h and sum the values. Unfortunately, there is no clear documentation on which model has which style of reporting. Further, some models switch styles at random times throughout the year. For this reason, we implement a check when downloading the models that tests which style of precipitation reporting is used. In particular, we test whether the difference from one hour to the next is always positive, which would be true if the total precipitation accumulation style was used. To further complicate things, as noted in the limited online documentation at \url{https://data.nssl.noaa.gov/thredds/fileServer/FRDD/HREF/2020/spc_href_data_guide.pdf}, some models sometimes use a two hour accumulation instead of a one hour accumulation. To detect the time windows for which this is the case, we check whether a given model's domain-averaged values are significantly ($\times 1.5$ or more) greater than that of a reference model in which we have confidence, ``nam\_conusnest". To check whether we did all the data transformations correctly, we compared numerous samples of our final HREF ensemble against the HREF Ensemble Viewer (\url{https://www.spc.noaa.gov/exper/href/}) to confirm that they matched.

\section{SUPPLEMENTARY DEFINITIONS}

 \noindent \textbf{Fractional Skill Score (FSS):} FSS \citep{roberts2008assessing}  is a deterministic metric for forecast accuracy which, unlike pixel-wise metrics like RMSE and MAE, avoids over-penalising small spatial shifts. For a given threshold, and a given window size, let $M_i$ be the fraction of pixels in window $i$ forecast to exceed the threshold, and let $O_i$ be the fraction of observations in window $i$ exceeding the threshold, then 
 \begin{align}
     FSS = 1 - \frac{\sum_{i} (O_i - M_i)^2}{\sum_{i}O_i^2 + \sum_{i}  M_i^2}.
 \end{align}
 We use an approximation of this metric in our initial training of the (deterministic) correction $g(0,x)$ (Stage 1) as a learning signal to encourage skillful low-res corrected forecasts. Specifically, instead of calculating $M_i$ and $O_i$ as the averages of binary masks $1_{y_i > c}$ over a given window for a threshold $c$,  we $M_i$ and $O_i$ are the averages of $\phi(10(y_i-c))$ for predictions and observations respectively, where $\phi$ is the sigmoid function.

\begin{figure}[h!]
    \centering
    \includegraphics[width=0.45\textwidth]{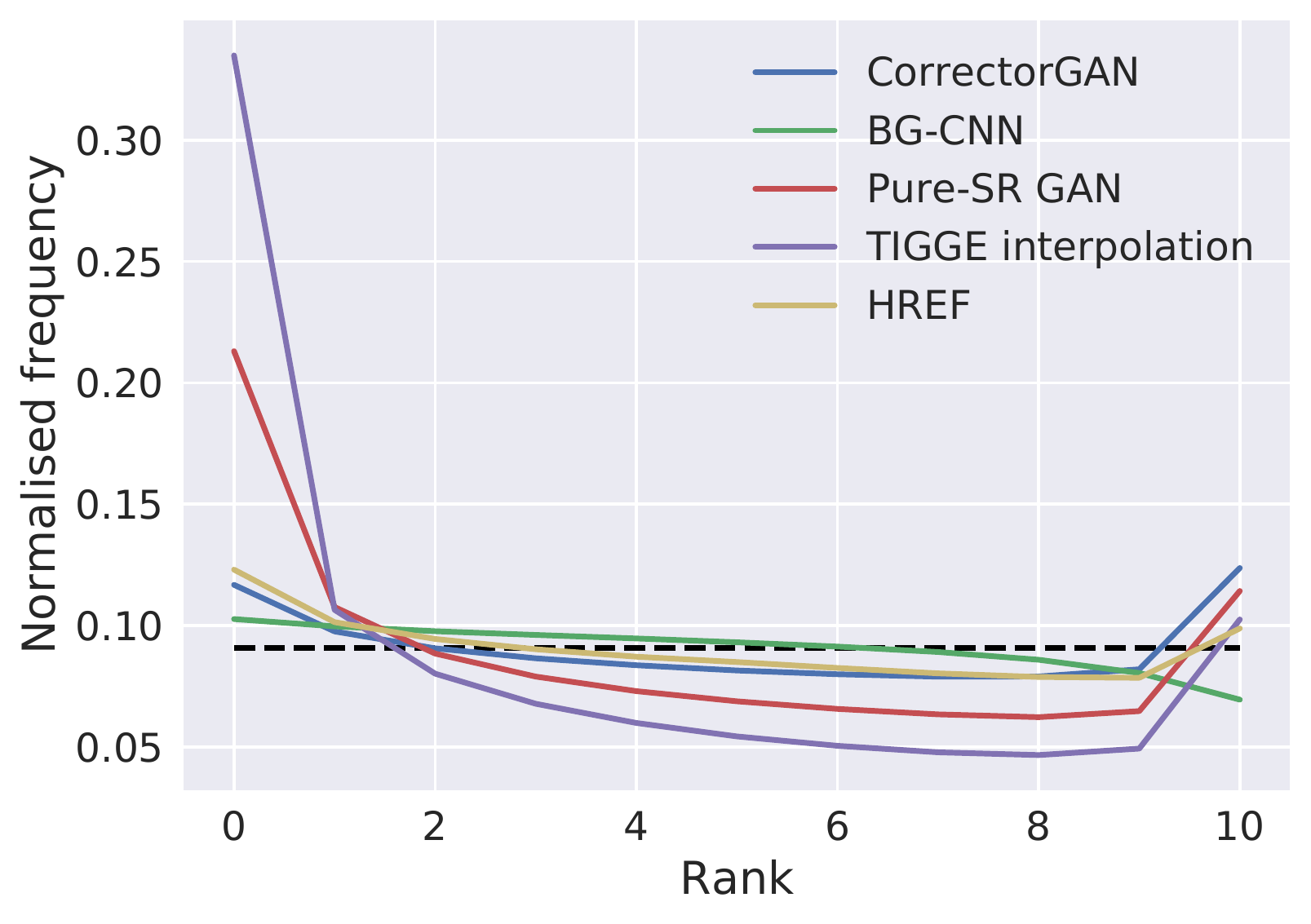}
    \caption{Rank Histogram}
    \label{fig:rankhist}
\end{figure}

\section{ADDITIONAL EVALUATION}

\subsection{Rank Histogram}
One of the most common approaches for evaluating whether an ensemble forecast fulfills the consistency desideratum (that is, whether observations $y_i$ behave like random draws from the generated forecast distribution) is to construct a rank histogram \citep{daniel2006statistical}. For each pixel in the entire test set, we record the rank (index) of the observed value when inserted into a sorted list of forecast ensemble members, and we then plat a histogram of these ranks. A perfectly consistent forecast would result in a uniform (flat) histogram. 

Figure $\ref{fig:rankhist}$ shows the rank histogram of CorrectorGAN, BG-CNN, Pure-SR GAN, HREF, and the TIGGE interpolation. Again we see similar results when comparing CorrectorGAN and HREF, both of which exhibit much flatter rank histograms than the TIGGE interpolation, which exhibits a well known problem of global, coarse forecasts sometimes known as a `drizzle bias': too often it predicts rain when there is none. The slightly larger peak by CorrectorGAN over HREF on the far right indicates that there are still some instances of heavy rainfall which are not predicted in the right pixels by the GAN, which are better localised by HREF. Pure-SR GAN exhibits some of the drizzle-bias seen in the TIGGE interpolation, which is to be expected given that it was not trained to correct such biases. BG-CNN presents a fairly flat rank-histogram but with a slight over-forecasting bias \citep{wilks2011statistical}.

\subsection{Full TIGGE Ensembles}
\begin{figure}[!h]
    \centering
    \includegraphics[width=\textwidth]{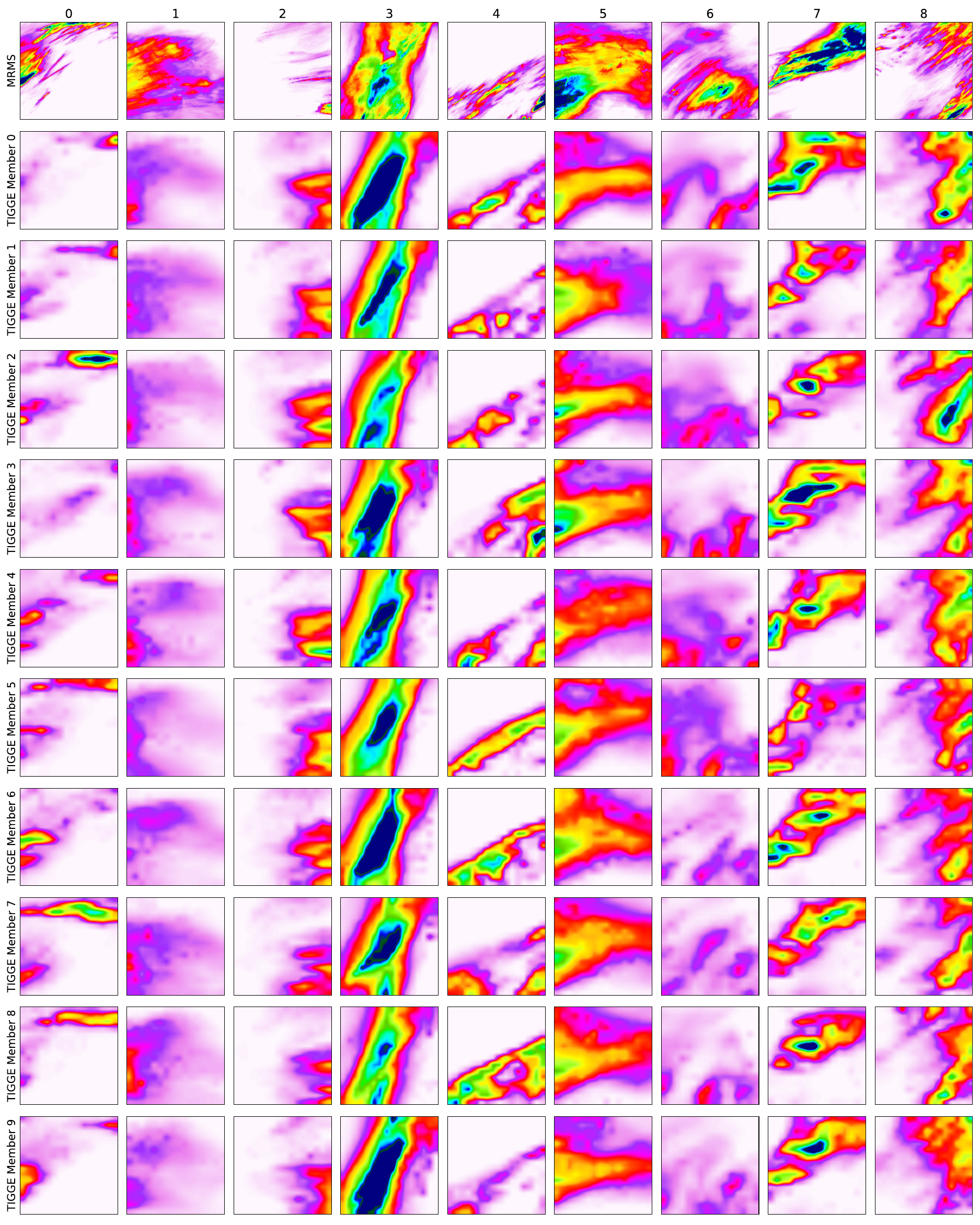}
    \caption{Full TIGGE ensembles for the geo-patches shown in Figure 3 in the paper.}
    \label{fig:full_tigge}
\end{figure}

Figure \ref{fig:full_tigge} shows the full TIGGE ensembles corresponding to the geo-patches shown in Figure 3 in the paper. Comparing with the full TIGGE ensemble confirms that CorrectorGAN does indeed achieve large scale correction in certain instances. 

\subsection{Deterministic metrics}
Table \ref{tab:deterministic} compares the 5 methods according to 4 deterministic metrics. We include these results for completeness, despite the fact that they are not the most appropriate for evaluating probabilistic forecasts. We compute the root-mean-squared error, and fractional skill score given a precipitation threshold of 4mm, for each ensemble member, for each test patch, and report the average for each method. We see that both CorrectorGAN and the TIGGE interpolation baseline perform very similarly in both metrics, outperforming all other methods on RMSE, and under-performing HREF on FSS. The significantly higher FSS achieved by BG-CNN represents a notable outlier given its ranking according to all other metrics. 

We also report precision and recall for binarised rainfall events at different thresholds. However, we emphasise that these metrics are not the most appropriate method of comparison. This is because in cases where the ground truth probability of extreme rainfall is low, as it often is, one would actually expect only a small fraction of realizations to have extreme rain (i.e. in these cases we would expect few `true positives' among the ensemble). Instead, we would like \textit{probability} assigned to such events to reflect the underlying probability - but without access to ground truth probabilities, the best we can do is evaluate how well the assigned probabilities calibrate with observed frequency of those events across the test dataset. This motivates the reliability diagrams as more informative for evaluation.

\begin{table}[]
\centering
\label{tab:deterministic}
\caption{Deterministic metrics}
\begin{tabular}{llcccccccccc}
  & Model                     & RMSE & FSS  & \multicolumn{4}{c}{Precision} & \multicolumn{4}{c}{Recall} \\ \cline{2-12} 
  &                           &  \    &  \    & 1mm   & 5mm   & 10mm  & 30mm  & 1mm   & 5mm  & 10mm & 30mm \\ \cline{5-12} 
0 & CorrectorGAN      & 1.557 & 0.61  & 0.65  & 0.57  & 0.44  & 0.16  & 0.59  & 0.41 & 0.24 & 0.02 \\
1 & BG-CNN           & 1.81 &  0.74 & 0.54  & 0.42  & 0.30  & 0.07  & 0.65  & 0.53 & 0.41 & 0.16 \\
2 & Pure-SR GAN       & 1.59 & 0.59 & 0.59  & 0.55  & 0.44  & 0.14  & 0.67  & 0.36 & 0.18 & 0.02 \\
3 & TIGGE interpolation & 1.557 & 0.62 & 0.63  & 0.56  & 0.46  & 0.20  & 0.68  & 0.46 & 0.29 & 0.06 \\
4 & HREF            & 1.6 & 0.63   & 0.63  & 0.51  & 0.37  & 0.11  & 0.57  & 0.45 & 0.35 & 0.17 \\ \cline{2-12} 
\end{tabular}


\end{table}

\section{ABLATION STUDY}
We study the importance both of the additional TIGGE variables (beyond precipitation) as well as the proposed staged training procedure with the following ablations. We compare CorrectorGAN trained as described above, with (1) \textit{CorrectorGAN (NPT)}: standard, end-to-end GAN training (i.e. no pre-training stages) of the CorrectorGAN architecture, with the full loss function specified in Stage 3 (Section \ref{sec:Model}), with the full 24-channel input, and (2) \textit{LeinGAN}: standard, end-to-end GAN training for a GAN model with 1-channel inputs comprising individual precipitation forecasts (i.e. removing all other TIGGE variables). This GAN model is essentially the same as the one proposed by \cite{leinonen2020stochastic}, but without the time-series consistency components, and is trained with their proposed loss. 

The CRPS and Brier Scores are shown in Table \ref{table:ablation-metrics}. The results across both metrics (and all thresholds) indicate that the \textit{Corrector + Super-resolver} design of our the CorrectorGAN generator, along with the addition of other TIGGE variables and ensemble members as inputs, brings improvements over and above LeinGAN. The addition of the staged training procedure induces further gains. 

\begin{figure}[h!]
    \centering
    \includegraphics[width = 0.9\textwidth]{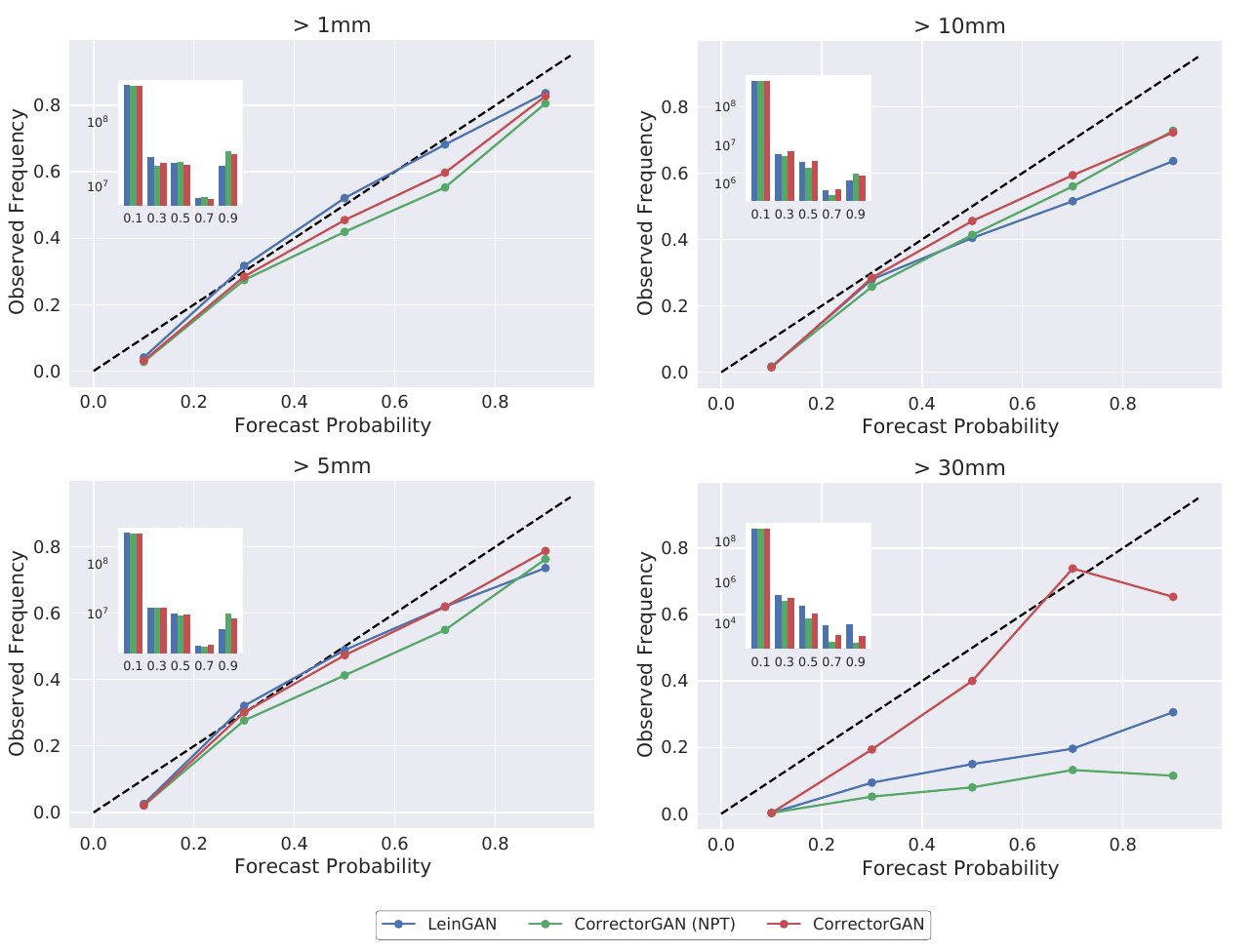}
    \caption{Ablation study reliability diagrams for 1mm, 5mm, 10mm, and 30mm precipitation thresholds. The insets show the number of predictions in each bin for each method (on a log scale).}
    \label{fig:ablation-reliability}
\end{figure}

The importance of these elements of CorrectorGAN is further corroborated by comparing their reliability diagrams, shown in Figure \ref{fig:ablation-reliability}. Except for the 1mm threshold, CorrectorGAN matches or exceeds the reliability of the other methods. However, unlike in Table \ref{table:ablation-metrics}, \textit{LeinGAN} mostly outperforms \textit{CorrectorGAN (NPT)} on reliability.

For completeness the deterministic metric results of the ablation study are shown in Table \ref{ablation-det-metrics}.

\begin{table*}[]
\centering
\label{table:ablation-metrics}
\caption{Ablation study probabilistic metrics.}
\begin{tabular}{lllllll}
  & Model                     & CRPS  & \multicolumn{4}{l}{Brier Score} \\ \cline{2-7} 
  &                           &       & 1mm    & 5mm   & 10mm  & 30mm   \\ \cline{4-7} 
0 & LeinGAN                   & 0.628 & 0.067  & 0.038 & 0.022 & 0.0025 \\
1 & Correctorgan (NPT) & 0.593 & 0.060  & 0.035 & 0.021 & 0.0025 \\
2 & CorrectorGAN              & 0.574 & 0.060  & 0.034 & 0.020 & 0.0024 \\ \cline{2-7} 
\end{tabular}
\end{table*}

\begin{table}[]
\centering
\label{ablation-det-metrics}
\caption{Ablation study deterministic metrics}
\begin{tabular}{llcccccccccc}
  & Model                     & RMSE & FSS  & \multicolumn{4}{c}{Precision} & \multicolumn{4}{c}{Recall} \\ \cline{2-12} 
  &                           &      &      & 1mm   & 5mm   & 10mm  & 30mm  & 1mm   & 5mm  & 10mm & 30mm \\ \cline{5-12} 
0 & LeinGAN                   & 1.66 & 0.50 & 0.62  & 0.51  & 0.40  & 0.10  & 0.48  & 0.32 & 0.19 & 0.02 \\
1 & CorrectorGAN (NPT) & 1.56 & 0.63 & 0.65  & 0.56  & 0.46  & 0.05  & 0.63  & 0.44 & 0.21 & 0.00 \\
2 & CorrectorGAN              & 1.56 & 0.61 & 0.65  & 0.57  & 0.44  & 0.16  & 0.59  & 0.41 & 0.24 & 0.02 \\ \cline{2-12} 
\end{tabular}
\end{table}
\end{document}